\def\year{2019}\relax
\documentclass[letterpaper]{article} %DO NOT CHANGE THIS
\usepackage{aaai19}  %Required
\usepackage{times}  %Required
\usepackage{helvet}  %Required
\usepackage{courier}  %Required
\usepackage{url}  %Required
\usepackage{graphicx}  %Required
\usepackage{amssymb,amsmath}
\usepackage{epsfig,epstopdf}
\usepackage{subfigure}
\usepackage{multirow}
\usepackage{float}
\usepackage{graphics}
\usepackage[ruled]{algorithm}
\usepackage{algorithmic}
\usepackage{comment}
\usepackage{placeins}
\usepackage{microtype}
\usepackage{balance}
\usepackage{xcolor}%required
\usepackage{soul}%required
\usepackage{bm}
\usepackage{enumerate}

\begin{document}
\title{Multi-View Multi-Instance Multi-Label Learning based on Collaborative Matrix Factorization\thanks{Corresponding author, gxyu@swu.edu.cn (Guoxian Yu).}}

\author{Yuying Xing$^1$, Guoxian Yu$^{1,2,*}$, Carlotta Domeniconi$^3$, Jun Wang$^1$, Zili Zhang$^{1,4}$ and Maozu Guo$^5$\\
$^1$College of Computer and Information Science, Southwest University, Chongqing, China \\
$^2$Hubei Key Laboratory of Intelligent Geo-Information Processing, China University of Geosciences, Wuhan, China\\
$^3$Department of Computer Science, George Mason University, Fairfax, USA\\
$^4$School of Information Technology, Deakin University, Geelong, Australia\\
$^5$School of Electrical and Information Engineering, Beijing University of Civil Engineering and Architecture, Beijing, China\\
\{yyxing4148, gxyu, kingjun, zhangzl\}@swu.edu.cn, carlotta@cs.gmu.edu, guomaozu@bucea.edu.cn}

\maketitle
\begin{abstract}
Multi-view Multi-instance Multi-label Learning(M3L) deals with complex objects encompassing diverse instances, represented with different feature views, and annotated with multiple labels.  Existing M3L solutions only partially explore the inter or intra relations between objects (or bags), instances, and labels, which can convey important contextual information for M3L. As such, they may have a compromised performance.

In this paper, we propose a  collaborative matrix factorization based solution called M3Lcmf. M3Lcmf first uses a heterogeneous network composed of nodes of bags,  instances, and labels, to encode different types of relations via multiple relational data matrices. To preserve the intrinsic structure of the data matrices, M3Lcmf collaboratively factorizes them into low-rank matrices, explores the latent relationships between bags, instances, and labels, and  selectively merges the data matrices. An aggregation scheme is further introduced to aggregate the instance-level labels into bag-level and to guide the factorization. An empirical study on benchmark datasets show that M3Lcmf outperforms other related competitive solutions both in the instance-level and bag-level prediction.
\end{abstract}

\section{Introduction}
Multi-Instance Multi-Label learning (MIML) is a  framework for modeling complex objects, in which each object (or bag) contains one or more instances and is annotated by several semantic labels \cite{Zhou2012MIML}. Let's consider $n$ bags $\mathcal{B}_i=\{\mathbf{x}_{i_1}, \mathbf{x}_{i_2}, \cdots, \mathbf{x}_{i_k}\}$ ($i=1,\dots,n$), where each bag encompasses $n_i \geq 1$ instances, and $\mathbf{x}_{i_j} \in \mathbb{R}^d$ is the feature vector of the $j$-th instance of the $i$-th bag. The $n$ bags and the $m=\sum_{i=1}^n n_i$ instances are annotated with
$q$ distinct labels.  $\mathbf{Y}_i\in \mathbb{R}^{1\times q}$ is the $q$-dimensional label vector for the $i$-th bag.
Given a training dataset $\mathcal{D}=\{(\mathcal{B}_{i},\mathbf{Y}_{i})\}_{i=1}^n$ , MIML aims at learning an instance-level $f(\mathbf{x})\in\mathbb{R}^{q}$ (or bag-level) predictor, which maps the input features of instances (or bags) onto the label space.

\begin{figure}[h!t]
\centering
 \includegraphics[width=0.42\textwidth]{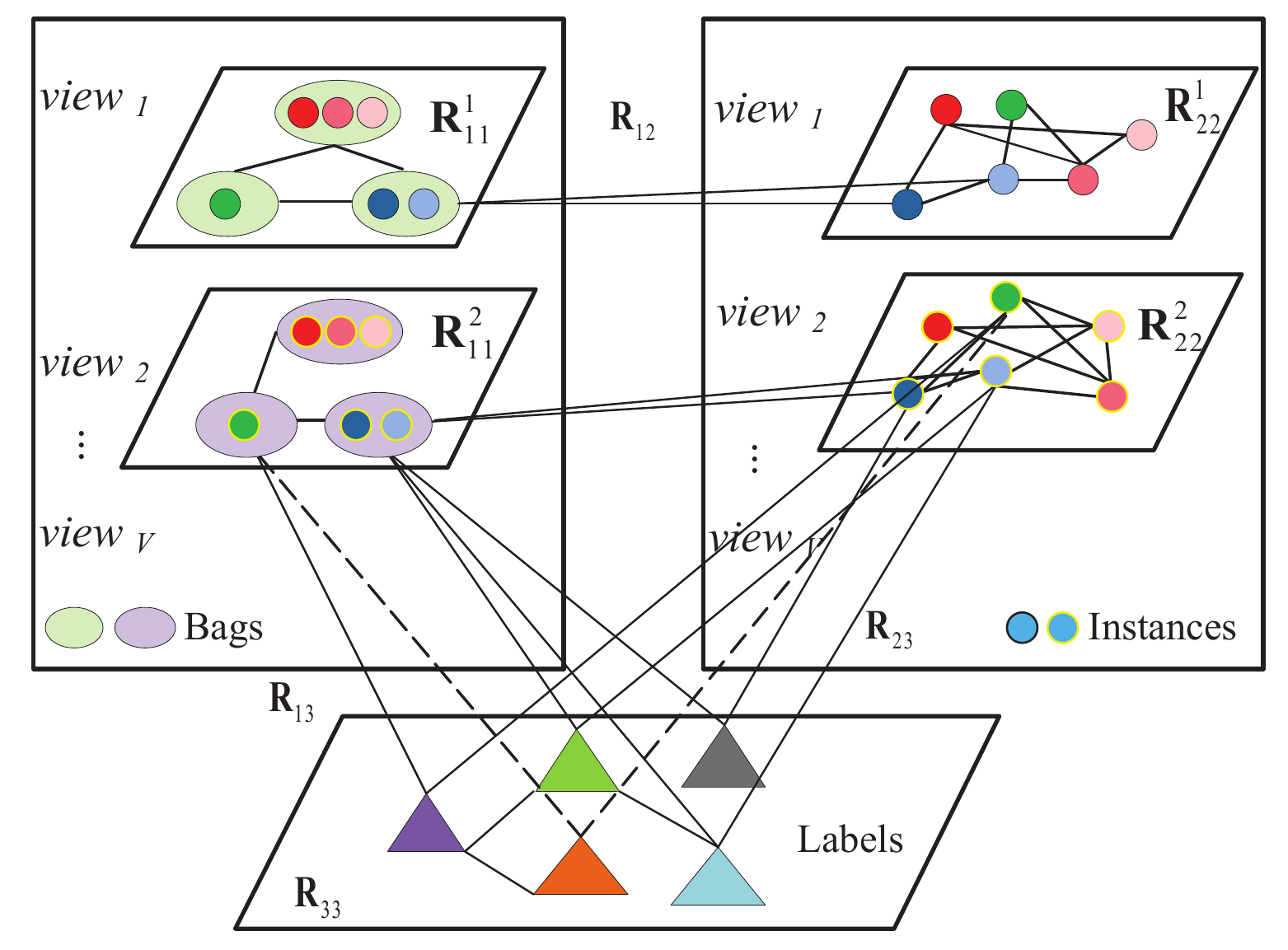}
 \vspace{-0.5em}
 \caption{An illustrative example of multi-view multi-instance multi-label objects. $\{\mathbf{R}^v_{ij}\}_{v=1}^V$ are the multi-type relational data matrices between bags (objects), instances, and labels across $V$ heterogeneous feature views.}
 \label{fig1}
 \vspace{-0.5em}
\end{figure}

Most MIML algorithms focus on single view data, where instances of bags are represented by one set of features. However, in real-world applications, a multi-instance multi-label object can often be represented via different views \cite{Nguyen2013Multi,shao2016multi}.
For example, as shown in Figure \ref{fig1}, three exemplar bags encompassing diverse instances are represented with $V$ heterogenous feature views. Since there are multi-type relations between bags and between instances, learning from multi-view bags is more difficult and challenging than the recently heavily studied MIML task  \cite{feng2017deep,zhu2017discover}.

Several Multi-view Multi-instance Multi-label Learning (M3L) approaches have been proposed to tackle this challenge \cite{Nguyen2013Multi,Nguyen2014Labeling,Li2017Multi,Yang2018KDD}.
\begin{table*}[h!tbp]
\centering
\scriptsize
\caption{Relations exploited by representative M3L and MIML methods.}
\begin{tabular}{|l|c c c c c c |}
\hline
 &\multicolumn{6}{c}{Relations}\\
 \cline{2-7}
   &bag-bag  &instance-instance &label-label &bag-instance &bag-label &instance-label\\
\hline
M3LDA\cite{Nguyen2013Multi}        & & &\checkmark &\checkmark &\checkmark &\checkmark \\
\hline
MIMLmix\cite{Nguyen2014Labeling}   & & &\checkmark &\checkmark &\checkmark &\checkmark \\
\hline
M3DN\cite{Yang2018KDD}             & & &\checkmark &\checkmark &\checkmark &\checkmark \\
\hline
M$^{2}$IL\cite{Li2017Multi}        & &\checkmark & &\checkmark &\checkmark & \\
\hline
MIMLSVM\cite{Zhou2008MIML}         &\checkmark  & & &\checkmark &\checkmark &\\
\hline
MIMLfast\cite{Huang2013Fast}       & & &\checkmark &\checkmark &\checkmark & \\
\hline
MIMLRBF\cite{Zhang2009MIMLRBF}     & & & &\checkmark &\checkmark &\checkmark \\
\hline
Proposed M3Lcmf                    &\checkmark &\checkmark &\checkmark &\checkmark &\checkmark &\checkmark \\
\hline
\end{tabular}
\label{table1}
\vspace{-0.5em}
\end{table*}
\citeauthor{Nguyen2013Multi} \shortcite{Nguyen2013Multi} pioneered an approach called M3LDA, which employs Latent Dirichlet Allocation \cite{Blei2003Latent} to explore the visual-label topics from the visual view and the text-label topics from the text view, and then enforces the predicted labels from the two respective views to be consistent. \citeauthor{Nguyen2014Labeling} \shortcite{Nguyen2014Labeling}  introduced another M3L approach, called MIMLmix, to leverage multiple views using a hierarchical Bayesian network and variational inference. MIMLmix can handle samples which are absent in some views. \citeauthor{Li2017Multi} \shortcite{Li2017Multi}  developed  a multi-view  multi-instance learning algorithm (M$^{2}$IL), which generates different graphs with different parameters to represent various contextual relations between instances of a bag. It then integrates these graphs into a unified framework for bag classification based on sparse representation and multi-view dictionary learning. \citeauthor{Yang2018KDD} \shortcite{Yang2018KDD} introduced a deep neural network based approach called M3DN. M3DN separately applies a deep network for each view, and requires the bag-based predictions from different views to be consistent within the same bag. In addition, M3DN adopts the Optimal Transport theory \cite{villani2008optimal} to capture the geometric information of the underlying label space and to quantify the quality of predictions.

However, these M3L approaches, like MIML solutions, only consider \emph{limited} types of relations between bags or between instances, as summarized in Table \ref{table1}. M3L approaches generally capture the relations between bags and instances, and the associations between bags and labels. Some approaches additionally exploit the relations between bags \cite{Zhou2008MIML}, between instances \cite{Li2017Multi}, and the correlations between labels \cite{Huang2013Fast,Yang2018KDD}. Furthermore,  other approaches use the associations between instances and labels \cite{Zhang2009MIMLRBF,Nguyen2014Labeling} to learn labels of bags at the instance level. All these types of relations simultaneously exist in M3L, however, \emph{none} of the existing solutions explicitly accounts for all these relations.

To take advantage of multiple feature views of instances (or bags), an intuitive solution is to concatenate features from different views into a long vector, and then to apply MIML algorithms on the concatenated vector. However, this concatenation causes over-fitting on a small number of training samples, and ignores the specific statistical property of each view \cite{xu2013survey}. Ensemble learning can also work on multi-view data and MIML classifiers are readily available for each view.  But the base classifiers are separately trained on individual views; as such, they may have a low performance given the insufficient information of each view and the neglect of complementary information across views.  Subspace learning-based approaches \cite{he2016online,tan2018incomplete} aim at obtaining a latent subspace shared by multiple views under the assumption that the input views are generated from a latent subspace. Latent subspace-based solutions may alleviate the issue of the ``curse of dimensionality'', but may neglect the intrinsic structure of individual views.  For multi-view data, the intrinsic structures of bags and instances may be different across views. Therefore, a competent M3L approach should account for \emph{multiple types of relations} between bags, instances and labels, and the intrinsic structures of different feature views.

In this paper, we introduce an approach called M3Lcmf. {M3Lcmf first constructs a heterogeneous network composed of nodes of bags, instances, and labels, to capture the \emph{intra-relations} between nodes of the same type,  \emph{inter-relations} between bags and instances, between bags and labels, and between instances and labels. To respect and employ the intrinsic structure of the subnetworks of the intra and inter-relations, it collaboratively factorizes the association matrices of the subnetworks into low-rank matrices to pursue the low-rank representation of the nodes and the latent relationships among them, and also to selectively integrate multiple feature views of bags and instances. M3Lcmf additionally introduces an aggregation term into the factorization objective, which not only can aggregate the instance-label associations into bag-level, but also can reversely guide the prediction of these associations. The main contributions of this work are summarized as follows:
\begin{enumerate}[(i)]
\item Unlike existing solutions that can only account for several types of
 relations between bags and instances, M3Lcmf can simultaneously take into account multiple types of relations between bags, instances, and labels.

\item Our proposed M3Lcmf can selectively combine multiple feature views of bags and instances, preserve multiple intrinsic intra- and inter-relations without mapping inter-relations into the homologous network of bags or instances. It can make predictions at the instance-level and automatically aggregate the predictions to the bag-level.

\item Experimental results on benchmark datasets show that M3Lcmf performs favorably against the recently proposed M3L approaches MIMLmix \cite{Nguyen2014Labeling} and M$^{2}$IL \cite{Li2017Multi}, and other representative MIML methods (including MIMLSVM \cite{Zhou2008MIML}, MIMLNN \cite{Zhou2012MIML}, MIMLRBF \cite{Zhang2009MIMLRBF} and MIMLfast \cite{Huang2013Fast}). M3Lcmf is also robust to a wide range of input parameters.
\end{enumerate}

\section{The Proposed Method}
\subsection{Problem Formulation}
Without loss of generality, we assume instances (or bags) have $V$ feature views,  $\mathcal{B}_i^v=\{\mathbf{x}^v_{i_1}, \mathbf{x}^v_{i_2}, \cdots, \mathbf{x}^v_{i_k}\}$, where $\mathbf{x}^v \in \mathbb{R}^{d_v}(v=1,2, \cdots, V)$ is the feature space of instances in the $v$-th view. $\mathbf{Y}_{i}\in \mathbb{R}^{1\times q}$ is the $q$-dimensional label space for the $i$-th bag across all the views. The task of M3L is to learn a predictive function $f(\{\mathcal{B}^v\}_{v=1}^V, \mathbf{Y}) \in\mathbb{R}^{q}$, which maps multiple input feature views onto the label space.

To address this task, we first construct a heterogeneous network to encode multiple types of relations between bags, instances, and labels. Next, we collaboratively factorize the relational data matrices of the heterogeneous network into low-rank matrices, and predict the instance-label association based on the respective low-rank matrices; we then aggregate the instance-level predictions onto bag-level. The following two subsections elaborate on the network construction and collaborative matrix factorization.

\subsection{Heterogeneous Network Construction}
As shown in Figure \ref{fig1}, there are three types of nodes in the heterogeneous network: bags, instances, and labels. Each type of nodes has a different intrinsic structure. Bags and instances can have multiple heterogeneous feature views, which often provide complementary information. We first construct a heterogeneous network to represent intrinsic structures between nodes of multiple information sources.

It is recognized that relations among instances in a bag convey important contextual information in multi-instance learning, and they influence the overall performance \cite{Li2017Multi}. To explore the intrinsic structure of  instances, we construct a subnetwork of instances for each feature view. For simplicity, we measure the relation between $\mathbf{x}_i^{v}$ and $\mathbf{x}_j^{v}$ in the $v$-th view using the Gaussian heat kernel $\mathbf{R}_{11}^{v}(i,j)=exp(-\frac{||\mathbf{x}_i^v-\mathbf{x}_j^v||_{F}^2}{\sigma^2})$, where $\sigma$ is the average Euclidean distance between all the $m$ instances of the $v$-th view.

In M3L, a bag contains one or more instances and has its own characteristics, which are different from those of instances.
Here, we construct a bag subnetwork to capture the contextual information of bags based on a composite Hausdorff distance for each view as follows:
\begin{eqnarray} \label{eq1}
\scriptsize
\mathbf{H}(i,j)=\frac{1}{3}\sum_{p\in\eta}\left\{
             \begin{array}{lr}
             \frac{\sum_{a\in{\mathcal{B}}_{i}^{v}}min_{b\in{\mathcal{B}}_{j}^{v}}d(a,b)+\sum_{b\in{\mathcal{B}}_{j}^{v}}min_{a\in{\mathcal{B}}_{i}^{v}}d(a,b)}{|{\mathcal{B}}_{i}^{v}|+|{\mathcal{B}}_{j}^{v}|}\\
             (p=avg\nonumber), &  \\
             \\
             max\{max_{a\in{\mathcal{B}}_{i}^{v}}min_{b\in{\mathcal{B}}_{j}^{v}}d(a,b),\\
             max_{b\in{\mathcal{B}}_{j}^{v}}min_{a\in{\mathcal{B}}_{i}^{v}}d(a,b)\}(p=max), \\
             \\
             min_{a\in{\mathcal{B}}_{i}^{v}}min_{b\in{\mathcal{B}}_{j}^{v}}d(a,b)(p=min)
             \end{array}
\right.
\end{eqnarray}
where $p\in\eta=\{`avg', `max', `min'\}$, $d(a,b)$ is the Euclidean distance between two instances ($a$ and $b$).  Then, we define $\mathbf{R}_{22}^{v}(i,j)=exp(-\frac{\mathbf{H}(i,j)}{\sigma_{H}^2})$ as the similarity between the $i$-th bag and $j$-th bag in the $v$-th view, and $\sigma_H$ is set to the average composite Hausdorff distance between all the bags of this view. These three types of Hausdorff distances are widely used in MIML\cite{Zhou2012MIML}. Different Hausdorff distances have different focuses. The minimal Hausdorff distance indicates the minimal distance between all instances of one bag and those of another bag; the maximal Hausdorff distance computes the  maximum distance between instances of a bag and the nearest instances of another bag; while the average Hausdorff distance takes into account more geometric relations between instances of two bags \cite{Zhang2009}. This composite similarity can integrate the merits of the Hausdorff distance metrics.

In M3L, each bag is simultaneously annotated with several semantic labels, and the labels are not mutually exclusive. Different pairs of labels may have different degrees of correlation. Label correlation can be leveraged to boost the performance multi-label learning \cite{zhang2014mlreview}.  To quantify label correlations, we adopt the widely used cosine similarity to construct a subnetwork of labels. Since instances and bags share the same label space, only one label subnetwork is constructed. Let $\mathbf{Y}(\cdot, c) \in \mathbb{R}^n$ store the distribution of label $c$ across all the bags. The  correlation between two labels $c_1$ and $c_2$ can be empirically estimated as follows:
\begin{equation} \label{eq2}
\mathbf{R}_{33}(c_{1},c_{2})=\frac{\mathbf{Y}(\cdot, c_{1})^T \mathbf{Y}(\cdot, c_{2})} {\parallel \mathbf{Y}(\cdot, c_{1})\parallel \parallel \mathbf{Y}(\cdot, c_{2})\parallel}
\end{equation}
The specific distance metrics used to construct the three types of intra-relations in the subnetworks have been chosen for their simplicity and wide applicability. Other distance metrics can be used as well.

There are three types of inter-relations between bags, instances, and labels. The bag-instance inter-relational data matrix $\mathbf{R}_{12} \in \mathbf{R}^{n\times m}$ can be specified based on the known bag-instance associations, which are readily available in multi-instance data. The bag-label relational matrix $\mathbf{R}_{13} \in \mathbf{R}^{n \times q}$ can be directly specified based on the known labels of bags. For the instance-label relational data matrix $\mathbf{R}_{23} \in \mathbf{R}^{m \times q}$, since the initial labels of instances are generally unknown in multi-instance learning, we initially set $\mathbf{R}_{23}=\mathbf{0}$. If the labels of instances are partially known, we can also specify $\mathbf{R}_{23}$ based on the known labels of instances.

By referring to Table \ref{table1}, we can say that the heterogeneous network can account for all types of relations between bags, instances, and labels.

\subsection{Collaborative Matrix Factorization}
To combine multiple intra-relational data matrices $\mathbf{R}_{11}^v$ and $\mathbf{R}_{22}^v$, we can project all the data matrices onto a composite instance-instance intra-relational data matrix, or onto a composite bag-bag intra-relational data matrix, and then make prediction on the composite relational data matrix. This projection idea has been used to integrate multiple inter-connected subnetworks \cite{gligorijevic2015methods}. However, this projection may enshroud the intrinsic structures of different relational data matrices and compromise the performance. \citeauthor{2015Zitnik} \shortcite{2015Zitnik} recently introduced a data fusion framework (DFMF) based on matrix factorization. This framework does not need to map a heterogeneous network into a small homologous network, and it can leverage and preserve the intrinsic structures of multiple relational data matrices. The objective function of this framework is as follows:
\begin{equation}
 \begin{aligned}
    min\underset{\mathbf{G}\geq 0}{\mathbf{Z}(\mathbf{G},\mathbf{S}})&= \sum_{\mathbf{R}_{ij}\in\mathcal{R}}
    ||\mathbf{R}_{ij}-\mathbf{G}_{i}\mathbf{S}_{ij}\mathbf{G}_{j}^{T}||_{F}^{2}\\
    &+\sum_{t=1}^{max_{i}t_{i}}tr(\mathbf{G}^{\mathbf{T}}\Theta^{(t)}\mathbf{G})
 \end{aligned}
\label{eq3}
\end{equation}
where $||\cdot||_{F}^{2}$ is the Frobenius norm. $\mathbf{R}_{ij}\in \mathbb{R}^{n_{i}\times n_{j}}, i,j\in\{1, 2, \cdots, N\}$ stores the inter-relation between the $i$-th object and the $j$-th object. $\mathbf{G}_{i}\in\mathbb{R}^{n_{i}\times d_{i}}$, $\mathbf{G}_{j}\in\mathbb{R}^{n_{j}\times d_{j}}$, $\mathbf{S}_{ij}\in\mathbb{R}^{d_{i}\times d_{j}}(d_{i}\ll n_{i}, d_{j}\ll n_{j})$,
$\mathbf{G}=diag(\mathbf{G}_{1},...,\mathbf{G}_{N})$ where $\mathbf{G}_{i}$ is the low rank representation of the $i$-th object type, and $N$ is the number of object types. Suppose the $i$-th type of objects has $t_{i}$ data sources, represented by $t_{i}$ constraint matrices $\{\Theta_{i}^{t} \in \mathbb{R}^{n_i \times n_i}\}_{t=1}^{t_{i}}(t\in \{1,...max_{i}t_{i}\})$. $\mathbf{\Theta}^{(t)}=diag(\mathbf{\Theta}_{1}^{(t)},...,\Theta_{N}^{(t)})$, which collectively stores all the block diagonal matrices.

Based on the constructed heterogeneous network, and for the non-negativity of the inter and intra-relational data matrices, we extend Eq. (\ref{eq3}) and define the objective function of M3Lcmf as follows:
\begin{equation} \label{eq4}
\begin{aligned}
min \underset{\mathbf{G}_{1}, \mathbf{G}_{2}, \mathbf{G}_{3}\geq 0}{\mathbf{Z}(\mathbf{G}_{1}, \mathbf{G}_{2}, \mathbf{G}_{3})}&= ||\mathbf{R}_{12}-\mathbf{G}_{1}\mathbf{G}_{2}^{T}||_{F}^{2}\\
&+||\mathbf{R}_{13}-\mathbf{G}_{1}\mathbf{G}_{3}^{T}||_{F}^{2}\\
&+||\mathbf{R}_{13}-\mathbf{\Lambda}\mathbf{R}_{12}\mathbf{G}_{2}\mathbf{G}_{3}^{T}||_{F}^{2}\\
&+MR(\mathbf{G})\\
\end{aligned}
\end{equation}
where $\mathbf{G}_{1}\in \mathbb{R}^{n\times d}$, $\mathbf{G}_{2}\in \mathbb{R}^{m\times d}$, and $\mathbf{G}_{3}\in \mathbb{R}^{q\times d}$ are the low rank representations of multiple bags, instances, and labels, respectively. M3Lcmf has two prediction objectives. The first one is to predict instance-label associations $\mathbf{R}_{23}$ by approximating it to $\mathbf{G}_2\mathbf{G}_3^T$. The other objective is to predict labels of bags by approximating $\mathbf{R}_{13}$ to $\mathbf{G}_1\mathbf{G}_3^T$.
Instead of approximating $\mathbf{R}_{13}$ by $\mathbf{G}_1\mathbf{G}_3^T$, we add an aggregation term $||\mathbf{R}_{13}-\mathbf{\Lambda}\mathbf{R}_{12}\mathbf{G}_{2}\mathbf{G}_{3}^{T}||_{F}^{2}$ into Eq. (\ref{eq4}) to aggregate label information of instances to their originating bags. $\mathbf{\Lambda}\in\mathbb{R}^{n\times n}$ is a diagonal matrix, and $\mathbf{\Lambda}(i,i)=1/n_i$. This aggregation term is also driven  by the  multi-instance learning principle that the labels of a bag depend on the labels of its instances. Note, this aggregation term can reversely guide the pursue of $\mathbf{G}_2$ and $\mathbf{G}_3$. As such, the labels of instance can also be learnt from those of bags. The last term $MR(\mathbf{G})$ is the manifold regularization \cite{belkin2006manifold} on $\mathbf{G}$.

The intra-relations between bags, instances, and labels carry important contextual information, whose usage can improve the overall performance. Since $\mathbf{G}_1$, $\mathbf{G}_2$, and $\mathbf{G}_3$ can be viewed as the latent low-dimensional representation of bags, instances, and labels, we follow the idea of manifold regularization to enforce two data points with a high intra-association value being nearby in the low-dimensional space, and formulate the last term in Eq. (\ref{eq4}) as below to use three types of intra-associations:
\begin{equation}
\begin{aligned}
MR(\mathbf{G})&=\sum_{v=1}^{V}\boldsymbol{\alpha}_{v}tr(\mathbf{G}_{1}^{T}({\mathbf{D}_{11}^v-\mathbf{R}}_{11}^{v})\mathbf{G}_{1})\\
&+\sum_{v=1}^{V}\boldsymbol{\beta}_{v}tr(\mathbf{G}_{2}^{T}(\mathbf{D}_{22}^v-{\mathbf{R}}_{22}^{v})\mathbf{G}_{2})\\
&+tr(\mathbf{G}_{3}^{T}(\mathbf{D}_{33}-\mathbf{R}_{33})\mathbf{G}_{3})+\lambda_{1}||\boldsymbol{\alpha}||_{F}^{2}+\lambda_{2}||\boldsymbol{\beta}||_{F}^{2}\\
&s.t. \sum_{v=1}^{V}\boldsymbol{\alpha}_{v}=1, \sum_{v=1}^{V}\boldsymbol{\beta}_{v}=1.
\end{aligned}
\label{eq5}
\end{equation}
where $\boldsymbol{\alpha}_v $ and $\boldsymbol{\beta}_v$ are two parameters to balance the importance of the $v$-th bag view and $v$-th instance view, respectively. $\mathbf{D}_{11}^{v}$ and $\mathbf{D}_{22}^{v}$ are two series of diagonal matrices, with each diagonal entry equal to the row sum of $\mathbf{R}_{11}^v$ and $\mathbf{R}_{22}^v$, respectively; $\mathbf{D}_{33}$ follows a similar definition. $tr(\mathbf{G}_{1}^{T}(\mathbf{D}_{11}^v-{\mathbf{R}}_{11}^{v})\mathbf{G}_{1})$ can be viewed as the smoothness loss on the $v$-th bag view. $\lambda_1 \geq 0$ and $\lambda_2 \geq 0$ are introduced to avoid selecting single view alone. If these two parameters are excluded, only $\mathbf{R}_{11}^v$ and  $\mathbf{R}_{22}^v$ with the smallest loss will be selected. Our empirical study shows that $\boldsymbol{\alpha}_v $ and $\boldsymbol{\beta}_v$ can indeed selectively integrate different views and reduce the impact of noisy views by assigning smaller or zero weights to them. We can see that DFMF equally treats all the relational matrices $\{\mathbf{R}^v_{ij}\}_{i,j=1}^{3}$, it does not differentiate the different degrees of relevance of $\{ \mathbf{R}_{11}^v \}_{v=1}^V$ and $\{\mathbf{R}_{22}^v\}_{v=1}^V$ toward the prediction task. Unlike DFMF, which simply reverses the sign of $\{\mathbf{R}_{i,i}^v\}_{v=1}^V (i\in\{1,2,3\})$ to fulfil $\mathbf{\Theta}^{(t)}$ in Eq. (\ref{eq3}),  M3Lcmf uses the graph Laplacian matrix to guide the approximation, and has a good geometric explanation.

From the above analysis, we can conclude that M3Lcmf can predict labels for complicated objects both at instance-level and bag-level, and can simultaneously preserve multi-type relations between bags and instances. Besides the aggregation term, another distinction between M3Lcmf and DFMF is that the former can selectively combine multiple intra-relational data matrices, whereas the latter equally treats all the relational data matrices. As such, M3Lcmf can reduce the impact of noisy (or irrelevant) intra-relational data matrices for the target prediction task.

Following the idea of standard nonnegative matrix factorization \cite{lee2001NMF} and Alternating Direction Method of Multipliers (ADMM), we alternatively optimizes one variable of $\mathbf{G}_1$, $\mathbf{G}_2$, $\mathbf{G}_3$, $\boldsymbol{\alpha}_v$ and $\boldsymbol{\beta}_v$ one time with other variables fixed. Due to page limit, the optimization procedures of these variables are provided in the Supplementary file.

We then use the optimized $\mathbf{G}_{2}$  and $\mathbf{G}_{3}$ to approximate $\mathbf{R}^*_{23}$ (instance-label association matrix) as follows:
\begin{equation} \label{eq16}
\mathbf{R}^*_{23}=\mathbf{G}_{2}\mathbf{G}_{3}^{T}
\end{equation}
To further map the labels of instances onto the corresponding bag, we approximate the bag-label association matrix $\mathbf{R}^*_{13} \in \mathbb{R}^{l \times q}$ as follows:
\begin{equation} \label{eq17}
\mathbf{R}^*_{13}=\mathbf{\Lambda}\mathbf{R}_{12}\mathbf{R}^*_{23}
\end{equation}
As such, M3Lcmf can make label prediction both at the instance and bag levels.

\section{Experiments}
\subsection{Experimental Setup}
We perform three experiments  to investigate the performance of the proposed M3Lcmf. In the first experiment, six representative and related approaches, including four MIML methods (MIMLSVM \cite{Zhou2008MIML}, MIMLRBF \cite{Zhang2009MIMLRBF}, MIMLNN \cite{Zhou2012MIML}, and MIMLfast \cite{Huang2013Fast}) and two M3L methods (MIMLmix \cite{Nguyen2014Labeling} and M$^{2}$IL\cite{Li2017Multi}) are compared against M3Lcmf on both the bag-level and instance-level prediction. In the second experiment, four variants of M3Lcmf are designed to quantify the contribution of different types of relations. The third experiment studies the parameter sensitivity of M3Lcmf.

Nine publicly available multi-instance multi-label datasets from different domains are used for the experiments. The details of the datasets are given  in Table \ref{table2}. The first five datasets are collected from \url{http://lamda.nju.edu.cn/CH.Data.ashx} and \url{http://github.com/hsoleimani/MLTM/tree/master/Data}. They only have the bag-level labels and are used for evaluating the bag-level predictions. The original Delicious dataset includes 12234 bags with 223285 instances; to avoid an excessively heavy computational load, we randomly selected 1000 bags with 17613 instances from Delicious for the experiments. The last four datasets have instance-level labels \cite{winn2005object,briggs2012rank}, they are used for instance-level prediction and evaluation \cite{huang2017multi,chen2018CMAL}.

\vspace{-1em}
\begin{table}[ht!bp]
\centering
\scriptsize
\caption{Statistics of night datasets used for the experiments. $bag$, $instance$, and $label$ are the number of bags, instances, and labels, respectively. avgBI is the average number of instances per bag, and avgBL is the average number of labels per bag.}
\begin{tabular}{l |r r r r r r}
\hline
Dataset & bag  &instance &label &avgBI &avgBL \\
\hline
Haloarcula\_marismortui   &304  &951  &234  &3.1 &3.2 \\
Geobacter\_sulfurreducens &379  &1214 &320  &3.2 &3.1 \\
Azotobacter\_vinelandii   &407  &1251 &340  &3.1 &4.0 \\
Pyrococcus\_furiosus      &425  &1321 &321  &3.1 &4.5 \\
Delicious                 &1000 &17613 &20  &17.6 &2.8 \\
\hline
Letter Frost    &144  &565    &26   &3.9 &3.6 \\
Letter Carroll  &166  &717    &26   &4.3 &3.9 \\
MSRC v2         &591  &1758   &23   &1.0 &2.5 \\
Birds           &548  &10232  &13   &18.7&2.1 \\
\hline
\end{tabular}
\label{table2}
\end{table}
To evaluate the effectiveness of M3Lcmf, four widely-used multi-label evaluation metrics are adopted, including Ranking Loss (\emph{RankLoss}), macro AUC (Area Under receiver operating Curve) (\emph{macroAUC}), Average Recall (\emph{AvgRecall}), and Average F1-score (\emph{AvgF1}). Due to space limitation, the formal definition of these metrics is omitted here but can be found in \cite{zhang2014mlreview,gibaja2015mltutorial}. The smaller the values of \emph{RankLoss}, the better the performance is. As such, to be consistent with the other evaluation metrics, we report \emph{1-RankLoss} instead. For the latter metrics, larger values are an indication of a better performance.

\subsection{Prediction Results at the Bag-Level}
\begin{table*}[ht!bp]
\scriptsize
    \caption{Results of bag-level prediction on different datasets. $\bullet$/$\circ$ indicates whether M3Lcmf is statistically (according to pairwise $t$-test at 95\% significance level) superior/inferior to the other method.}
    \vspace{-1em}
     \begin{center}
    \begin{tabular}{c|c c c c|c c|c}
    \hline
     {Metric} &MIMLNN &MIMLRBF &MIMLSVM &MIMLfast &MIMLmix  &M$^2$IL &M3Lcmf\\
    \hline
    & \multicolumn{7}{c}{Haloarcula\_marismortui}\\
     \cline{2-8}
    \emph{1-RankLoss} &$0.713\pm0.029\bullet$   &$0.761\pm0.021\circ$    &$0.689\pm0.027\bullet$ &$0.553\pm0.022\bullet$ &$0.782\pm0.000\circ$   &$0.828\pm0.000\circ$   &$0.728\pm0.026$ \\
    \emph{macroAUC}   &$0.624\pm0.029\circ$     &$0.658\pm0.034\circ$    &$0.603\pm0.022\circ$   &$0.717\pm0.029\circ$   &$0.547\pm0.000\bullet$ &$0.442\pm0.000\bullet$ &$0.582\pm0.022$ \\
    \emph{AvgRecall}  &$0.079\pm0.015\bullet$   &$0.184\pm0.028\bullet$  &$0.175\pm0.022\bullet$ &$0.007\pm0.023\bullet$ &$0.002\pm0.000\bullet$ &$0.016\pm0.000\bullet$ &$0.299\pm0.041$ \\
    \emph{AvgF1}      &$0.128\pm0.019\bullet$   &$0.257\pm0.027\bullet$  &$0.218\pm0.022\bullet$ &$0.092\pm0.022\bullet$ &$0.033\pm0.000\bullet$ &$0.019\pm0.000\bullet$ &$0.301\pm0.022$ \\
    \hline
    & \multicolumn{7}{c}{Azotobacter\_vinelandii} \\
     \cline{2-8}
    \emph{1-RankLoss} &$0.656\pm0.021\bullet$   &$0.693\pm0.032\circ$    &$0.681\pm0.016\circ$   &$0.537\pm0.021\bullet$ &$0.813\pm0.000\circ$   &$0.805\pm0.000\circ$   &$0.663\pm0.019$ \\
    \emph{macroAUC}   &$0.564\pm0.048\bullet$   &$0.638\pm0.040\circ$    &$0.565\pm0.028\bullet$ &$0.666\pm0.021\circ$   &$0.621\pm0.000\circ$   &$0.509\pm0.000\bullet$ &$0.617\pm0.045$ \\
    \emph{AvgRecall}  &$0.069\pm0.024\bullet$   &$0.105\pm0.024\bullet$  &$0.116\pm0.021\bullet$ &$0.054\pm0.018\bullet$ &$0.019\pm0.000\bullet$ &$0.004\pm0.000\bullet$ &$0.178\pm0.022$ \\
    \emph{AvgF1}      &$0.109\pm0.033\bullet$   &$0.157\pm0.029\bullet$  &$0.148\pm0.023\bullet$ &$0.069\pm0.017\bullet$ &$0.072\pm0.000\bullet$ &$0.007\pm0.000\bullet$ &$0.199\pm0.013$ \\
    \hline
    & \multicolumn{7}{c}{Geobacter\_sulfurreducens} \\
     \cline{2-8}
    \emph{1-RankLoss} &$0.656\pm0.018\bullet$   &$0.688\pm0.024\circ$    &$0.694\pm0.020\circ$   &$0.552\pm0.019\bullet$ &$0.798\pm0.000\circ$   &$0.821\pm0.000\circ$   &$0.684\pm0.000$ \\
    \emph{macroAUC}   &$0.564\pm0.027\bullet$   &$0.608\pm0.033\circ$    &$0.567\pm0.015\circ$   &$0.691\pm0.022\circ$   &$0.375\pm0.000\bullet$ &$0.499\pm0.000\bullet$ &$0.567\pm0.000$ \\
    \emph{AvgRecall}  &$0.077\pm0.016\bullet$   &$0.129\pm0.021\bullet$  &$0.137\pm0.018\bullet$ &$0.042\pm0.009\bullet$ &$0.032\pm0.000\bullet$ &$0.012\pm0.000\bullet$ &$0.296\pm0.000$ \\
    \emph{AvgF1}      &$0.120\pm0.021\bullet$   &$0.186\pm0.026\bullet$  &$0.173\pm0.022\bullet$ &$0.058\pm0.009\bullet$ &$0.040\pm0.000\bullet$ &$0.014\pm0.000\bullet$ &$0.277\pm0.000$ \\
    \hline
    & \multicolumn{7}{c}{Pyrococcus\_furiosus} \\
     \cline{2-8}
    \emph{1-RankLoss} &$0.722\pm0.014\bullet$  &$0.732\pm0.000\bullet$   &$0.727\pm0.027\bullet$ &$0.469\pm0.035\bullet$ &$0.760\pm0.000\circ$   &$0.809\pm0.000\circ$   &$0.733\pm0.015$ \\
    \emph{macroAUC}   &$0.593\pm0.029\circ$    &$0.520\pm0.000\bullet$   &$0.613\pm0.043\circ$   &$0.469\pm0.030\bullet$ &$0.488\pm0.000\bullet$ &$0.485\pm0.000\bullet$ &$0.543\pm0.011$ \\
    \emph{AvgRecall}  &$0.069\pm0.017\bullet$  &$0.105\pm0.000\bullet$   &$0.134\pm0.029\bullet$ &$0.119\pm0.038\bullet$ &$0.004\pm0.000\bullet$ &$0.006\pm0.000\bullet$ &$0.341\pm0.038$ \\
    \emph{AvgF1}      &$0.086\pm0.015\bullet$  &$0.116\pm0.000\bullet$   &$0.174\pm0.034\bullet$ &$0.115\pm0.021\bullet$ &$0.056\pm0.000\bullet$ &$0.008\pm0.000\bullet$ &$0.307\pm0.025$ \\
    \hline
    & \multicolumn{7}{c}{Delicious} \\
     \cline{2-8}
    \emph{1-RankLoss} &$0.685\pm0.012\circ$  &$0.735\pm0.008\circ$   &$0.580\pm0.053\bullet$ &$0.466\pm0.023\bullet$ &$--$ &$0.439\pm0.000\bullet$  &$0.636\pm0.000$ \\
    \emph{macroAUC}   &$0.627\pm0.010\circ$  &$0.670\pm0.012\circ$   &$0.583\pm0.009\circ$ &$0.466\pm0.024\bullet$ &$--$   &$0.549\pm0.000\circ$  &$0.480\pm0.000$ \\
    \emph{AvgRecall}  &$0.112\pm0.014\bullet$&$0.029\pm0.019\bullet$ &$0.142\pm0.030\bullet$ &$0.619\pm0.045\circ$ &$--$   &$0.097\pm0.000\bullet$  &$0.178\pm0.000$ \\
    \emph{AvgF1}      &$0.180\pm0.018\bullet$  &$0.054\pm0.033\bullet$ &$0.201\pm0.032\bullet$ &$0.264\pm0.013\circ$ &$--$ &$0.136\pm0.000\bullet$  &$0.252\pm0.000$ \\
    \hline
    \end{tabular}
    \end{center}
    \label{table3}
\end{table*}

We randomly partition the samples of each dataset into a training set (70\%) and a testing set (30\%), and independently run each  algorithm in each partition. We report the average results (10 random partitions) and standard deviations in Table \ref{table3}. Since there are no off-the-shelf multi-view datasets for multi-instance multi-label learning,  for MIMLmix \cite{Nguyen2014Labeling}, M$^{2}$IL\cite{Li2017Multi} and the proposed M3Lcmf, we divide the original features of each bag into two views by randomly selecting half features for one view, and the remaining features for the other view. We initialize $\mathbf{R}_{12}(i,k)=1$ when the $i$-th bag encompasses the $k$-th instance; $\mathbf{R}_{12}(i,k)=0$ otherwise. We set $\mathbf{R}_{13}(i,c)=1$ when the $i$-th bag is annotated with the $c$-th label; $\mathbf{R}_{13}(i,c)=0$ otherwise. Both $\lambda_{1}$ and $\lambda_{2}$ are fixed to 1000, and the low-rank size of $\mathbf{G}_i$ ($i \in \{1,2,3\}$) is fixed to 140. The input parameters of these comparing methods are specified (or optimized) as suggested by the authors in their code or papers, and the setting of the parameters for M3Lcmf will be investigated later.

M3Lcmf generally outperforms these comparing methods across different datasets and the used metrics. We further used the signed-rank test \cite{demvsar2006statistical} to check the significance between M3Lcmf and these methods (except MIMLRBF). All the $p$-values are small than 0.02, and the $p$-value between M3Lcmf and MIMLRBF is 0.13. MIMLmix did not complete the computation on the Delicious dataset over the period of two weeks. As a result, we could not report the results of MIMLmix on this dataset. M3Lcmf, MIMLmix, and M$^{2}$IL are M3L methods, and M3Lcmf frequently outperforms the latter two, which only use limited types of relations between objects.
This fact shows the importance of accounting for multi-type relations in M3L. M3Lcmf has a lower \emph{1-RankLoss} but a higher \emph{AvgRecall} and \emph{AvgF1} than MIMLmix, the possible reason is that MIMLmix captures label correlations by assuming the labels being sampled from Multinomial distribution and it samples a label indicator for each instance, whereas M3Lcmf simply uses the cosine similarity to measure the correlation. M3Lcmf outperforms three  MIML solutions (MIMLNN, MIMLfast and MIMLSVM), which utilize much fewer relations between bags, instances and labels than M3Lcmf does. This comparison again corroborates the advantage of leveraging multiple types of relations in M3L, and also suggests the importance of integrating multiple data views. Although MIMLRBF considers limited types of relations between bags and instances, it still obtains a comparable performance with M3Lcmf. The possible cause is that MIMLRBF additionally uses the RBF neural network to learn an enhanced feature representation and a nonlinear classifier.

\subsection{Prediction Results at the Instance-Level}
\begin{table}[t]
\scriptsize
    \caption{Results on different multi-instance datasets. $\bullet$/$\circ$ indicates whether M3Lcmf is statistically (according to pairwise $t$-test at 95\% significance level) superior/inferior to the other methods.}
  \begin{center}
    \begin{tabular}{c|c c|c}
    \hline
     {Metric}  &MIMLfast &MIMLmix &M3Lcmf\\
    \hline
    & \multicolumn{3}{c}{Letter Frost} \\
     \cline{2-4}
     \emph{1-RankLoss}&$0.426\pm0.049\bullet$  &$0.667\pm0.000\bullet$ &$0.734\pm0.050$\\
     \emph{AvgF1}     &$0.094\pm0.015\bullet$  &$0.150\pm0.000\bullet$ &$0.352\pm0.107$\\
    \hline
    & \multicolumn{3}{c}{Letter Carroll} \\
     \cline{2-4}
     \emph{1-RankLoss}&$0.458\pm0.065\bullet$  &$0.410\pm0.000\bullet$ &$0.692\pm0.012$\\
    \emph{AvgF1}      &$0.096\pm0.023\bullet$  &$0.086\pm0.000\bullet$ &$0.104\pm0.012$\\
    \hline
    & \multicolumn{3}{c}{MSRC v2} \\
     \cline{2-4}
     \emph{1-RankLoss}&$0.419\pm0.030\bullet$ &$0.579\pm0.000\bullet$ &$0.652\pm0.005$\\
    \emph{AvgF1}      &$0.111\pm0.005\bullet$ &$0.333\pm0.000\circ$   &$0.208\pm0.074$\\
    \hline
    & \multicolumn{3}{c}{Birds} \\
     \cline{2-4}
    \emph{1-RankLoss} &$0.524\pm0.184\bullet$ &$0.937\pm0.000\circ$ &$0.666\pm0.000$\\
    \emph{AvgF1}      &$0.061\pm0.070\bullet$ &$0.503\pm0.000\circ$ &$0.286\pm0.000$\\
    \hline
    \end{tabular}
    \end{center}
    \label{table5}
    \vspace{-1em}
\end{table}
To investigate the performance of M3Lcmf at the instance-level, we conduct experiments on the last four datasets with instance-level labels in Table \ref{table5}. MIMLfast, MIMLmix and the proposed M3Lcmf are tested on these datasets under the same experimental protocol at the bag-level. The result values of \emph{1-RankLoss} and \emph{AvgF1} are reported in Table \ref{table5}.

M3Lcmf outperforms these comparing methods on different datasets in most cases, and it loses to MIMLmix on the Birds dataset. Among these three comparing methods, MIMLmix often ranks the 2nd place and MIMLfast the 3rd place. MIMLmix does not make use of bag-bag relation and instance-instance relation as summarized in Table \ref{table1}. MIMLfast additionally does not make use of instance-label relation, so it loses to MIMLmix, and say nothing of M3Lcmf, which utilizes all six types of relations. These comparisons again prove the effectiveness of leveraging multi-type relations in M3L. In summary, M3Lcmf can not only accurately predict labels of bags, but also labels of instances.

\subsection{Contribution of Different Types of Relations}
To further analyze the contribution of different relations used by M3Lcmf, we introduce four variants. (i) M3Lcmf (nR11) does not consider the relation between bags, i.e., $\mathbf{R}_{11}^{v}=0$; (ii) M3Lcmf (nR22) does not consider the relation between instances, i.e., $\mathbf{R}_{22}^{v}=0$; (iii) M3Lcmf (nR33) does not consider the relation between labels, i.e., $\mathbf{R}_{33}=0$; (iv) M3Lcmf (nR23) does not consider the relation between instances and labels, i.e., $\mathbf{R}^*_{13}=\mathbf{G}_{1}\mathbf{G}_{3}^{T}$, instead of $\mathbf{R}^*_{13}=\mathbf{\Lambda}\mathbf{R}_{12}\mathbf{G}_{2}\mathbf{G}_{3}^{T}$. We follow the experimental protocol at the bag-level prediction, and report the results  of \emph{1-RankLoss} obtained by M3Lcmf and its variants in Fig. \ref{fig2}.
\begin{figure}[h!t]
 \vspace{-0.5em}
\centering
\includegraphics[scale=0.3]{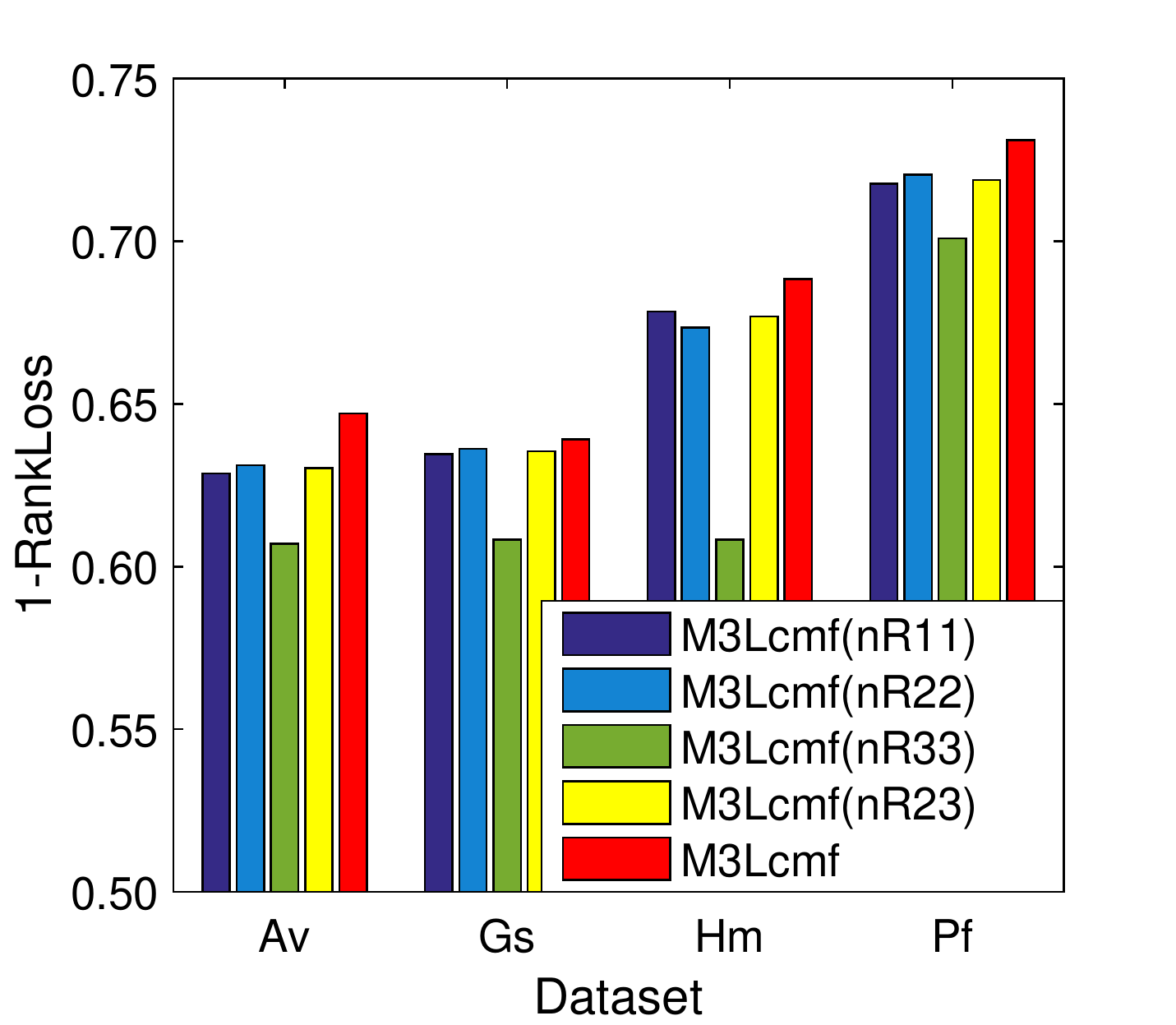}
 \vspace{-0.5em}
 \caption{\emph{1-RankLoss} of M3Lcmf and its variants on different datasets. Av: Azotobacter\_vinelandii, Gs: Geobacter\_sulfurreducens, Hm: Haloarcula\_marismortui, Pf: pyrococcus\_furiosus.}
 \label{fig2}
 \vspace{-0.5em}
\end{figure}

M3Lcmf significantly outperforms its variants, which separately disregard one type of relations.  M3Lcmf often outperforms M3Lcmf (nR11) and M3Lcmf (nR22). This observation suggests the relation between bags and that between instances have an important effect on M3Lcmf.
Besides, M3Lcmf(nR33) is outperformed by all the other variants, which shows the importance of considering the label correlation. In addition, we can observe that M3Lcmf (nR23) is outperformed by
M3Lcmf. This observation not only proves the effectiveness of the introduced aggregation term, but also shows the importance of instance-label relations in boosting the prediction performance.

From these results, we can conclude that multiple types of relations between bags, instances, and labels should be simultaneously considered in M3L.

\subsubsection{Parameter Sensitivity}
Three parameters ($\lambda_{1}$, $\lambda_{2}$, and the low-rank size $d$ of $\mathbf{G}$) may affect the performance of M3Lcmf. We conduct additional experiments to investigate the sensitivity of these parameters. For brevity, we only report the results on Azotobacter vinelandii and MSRC v2, and the results on the other datasets lead to similar conclusions.
\begin{figure}[h!t]
\centering
 \vspace{-1em}
\subfigure[\emph{Azotobacter vinelandii}]{\label{fig3a}\includegraphics[scale=0.25]{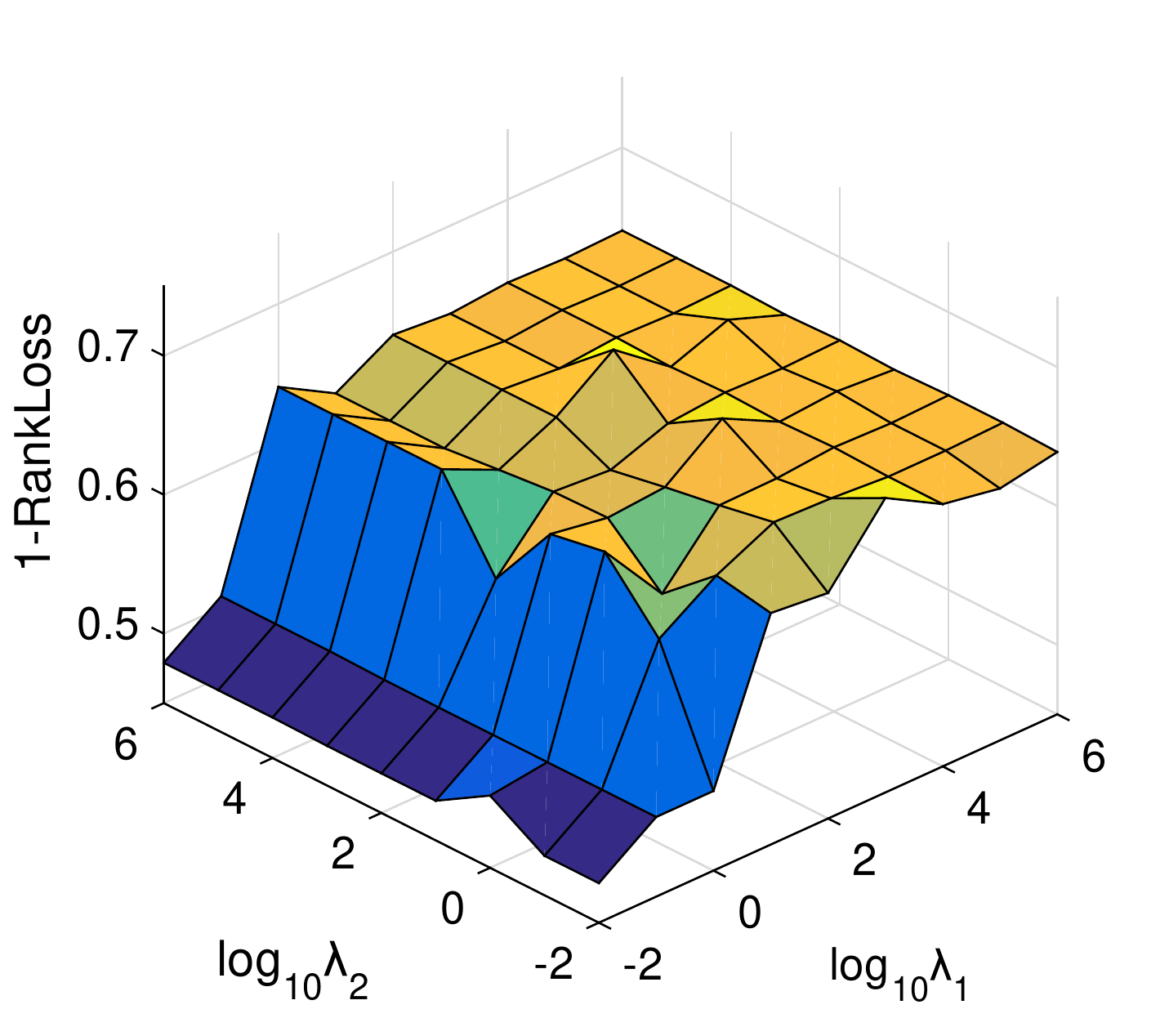}}
\subfigure[\emph{MSRC v2}]{\label{fig3b}\includegraphics[scale=0.21]{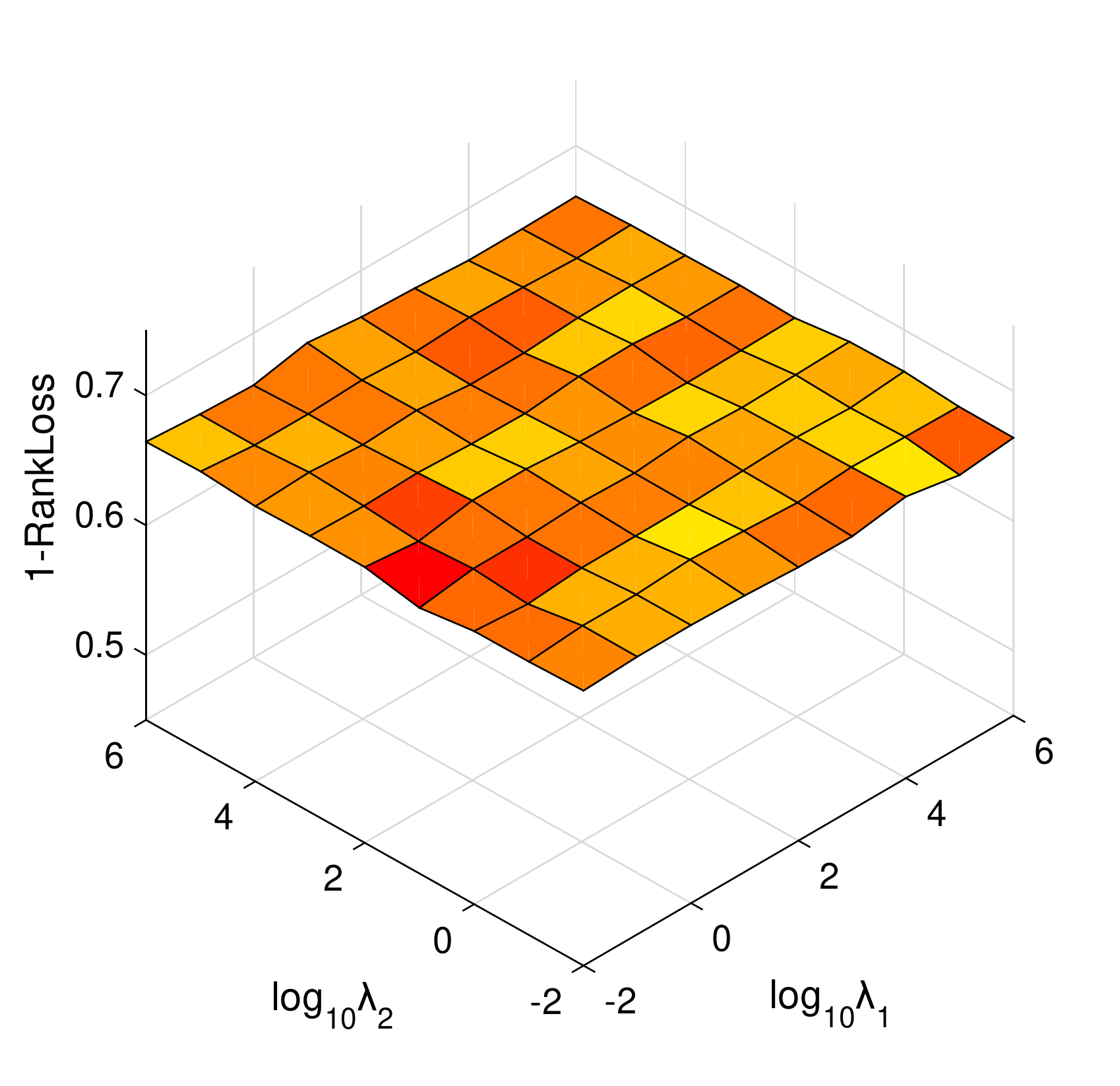}}
 \vspace{-0.5em}
 \caption{\emph{1-RankLoss} of M3Lcmf under different combinations of $\lambda_{1}$ and $\lambda_{2}$ on Azotobacter vinelandii and MSRC v2.}
  \vspace{-0.5em}
 \label{fig3}
\end{figure}

From the explicit solution for $\boldsymbol{\alpha}_v$ and $\boldsymbol{\beta}_v$ in the supplementary file, it is clear that once the values $\lambda_{1}$ and $\lambda_{2}$ are specified, the weights assigned to $\mathbf{R}^v_{11}$ and $\mathbf{R}^v_{22}$ can be computed based on the reconstruction loss of those matrices.
To investigate the sensitivity of these two parameters, we vary $\lambda_{1}$ and $\lambda_{2}$ in the range $\{10^{-2}, 10^{-1}, \cdots, 10^{6}\}$, and report the average \emph{1-RankLoss} of M3Lcmf under different combinations of them in Fig. \ref{fig3}. We can see that M3Lcmf achieves a stable  performance under a wide range of combinations of values for $\lambda_{1}$ and $\lambda_{2}$. For Azotobacter vinelandii, M3Lcmf achieves a good performance with $\lambda_{1}$ and $\lambda_{2}$ in $[10^{2},10^{6}]$, and it shows a significantly reduced \emph{1-RankLoss} when either $\lambda_1$ or $\lambda_2$ are set to a too small value. This is because the predictions are made and evaluated at the bag-level and the bag-level intra-relation plays a more important role, but only one bag-level intra relational data matrix is selected under this setting. Unlike the pattern on Azotobacter vinelandii, M3Lcmf holds a relatively stable performance on MSRC v2 under different combinations of values for $\lambda_{1}$ and $\lambda_{2}$. This is because Azotobacter vinelandii provides more structural information and feature information for the intra-relational data matrices of bags (or instances) than MSRC v2. Particularly, the former has more instances per bag than the latter, and the bag in MSRC v2 generally has one instance. Besides, the feature dimensionality of instances in Azotobacter vinelandii is much larger than that of MSRC v2. This investigation suggests the importance of structural information of bags (or instances) in M3L. From these results, we can conclude that an effective combination of $\lambda_{1}$ and $\lambda_{2}$ can be easily found.

\begin{figure}[h!t]
\centering
\vspace{-1em}
\subfigure[\emph{Azotobacter vinelandii}]{\label{fig4a}\includegraphics[scale=0.23]{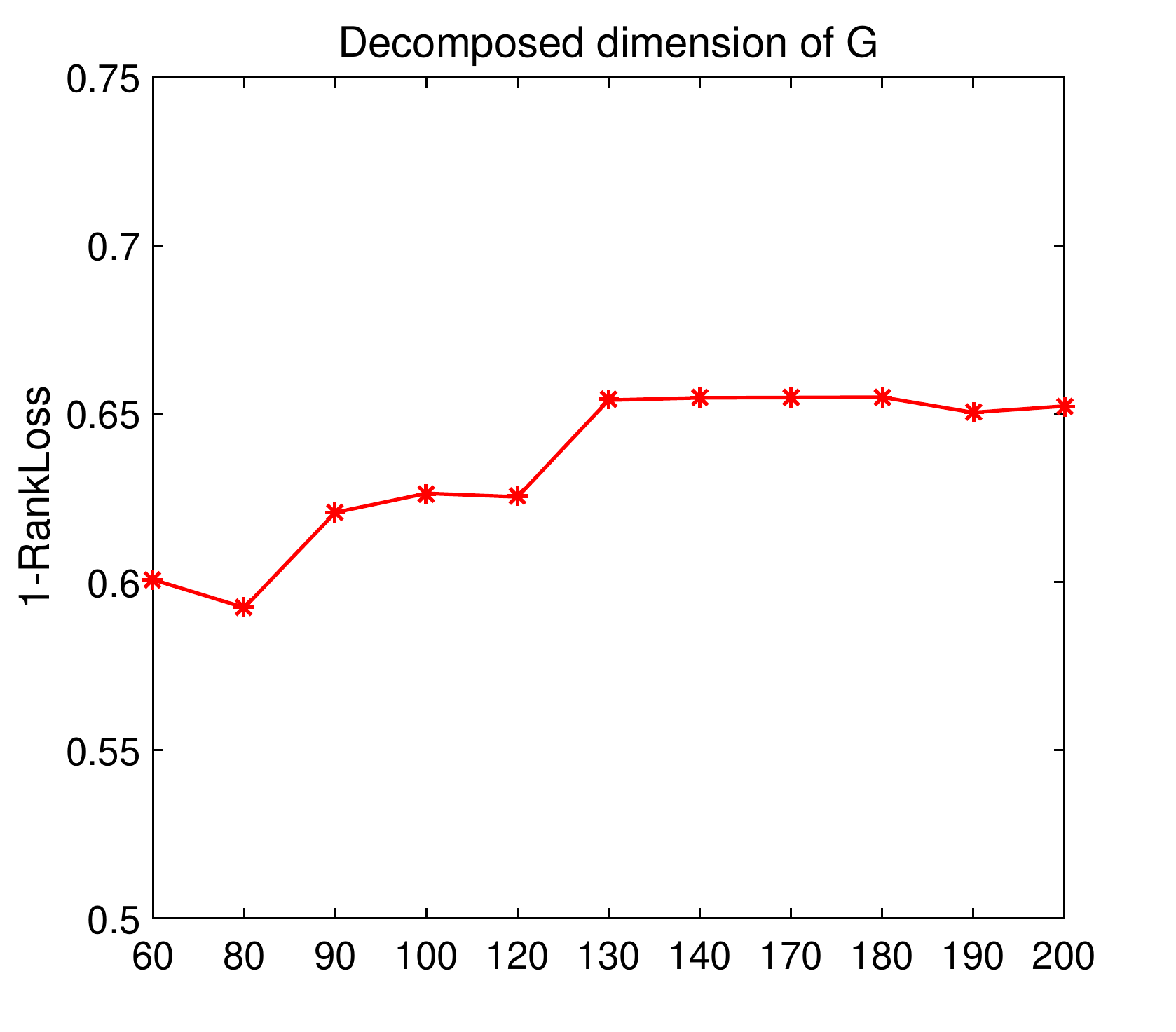}}
\subfigure[\emph{MSRC v2}]{\label{fig4b}\includegraphics[scale=0.28]{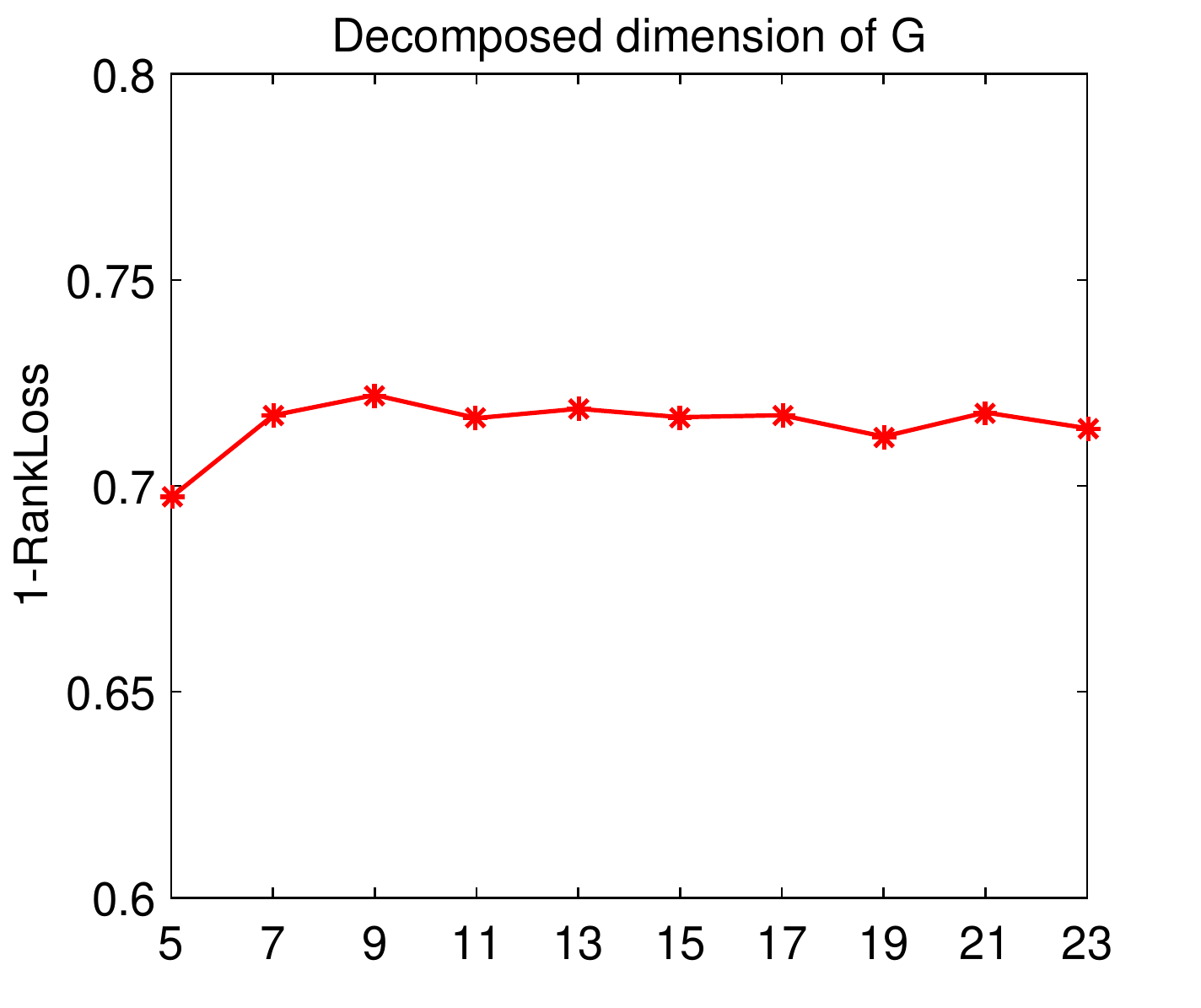}}
 \vspace{-0.5em}
 \caption{\emph{1-RankLoss} vs. $d$ (low-rank size) on Azotobacter vinelandii and MSRC v2.}
 \label{fig4}
 \vspace{-1em}
\end{figure}
The low-rank size $d$ of $\mathbf{G}$ is an essential parameter for M3Lcmf. Fig. \ref{fig4} shows the results of M3Lcmf under different input values of $d$ on Azotobacter vinelandii and MSRC v2 with $\lambda_{1}=10^3$ and $\lambda_{2}=10^3$. We observe an increasing trend of \emph{1-RankLoss}, and an overall good performance when $d\geq 140$ or $d\geq 11$. M3Lcmf does not show a high \emph{1-RankLoss} when a small $d$ is adopted, that is because a too small $d$ can not sufficiently encode the latent feature information of bags, instances, and labels. However, we can still find that an effective input value $d$ can be easily selected.

\subsection{Contributions of Weighting Intra-Relational Data}
To investigate the contribution of weighting intra-relational data and the capability of M3Lcmf on discarding noisy intra-relational data matrices, we added 10 synthetic \emph{noisy} intra-relational data matrices of bags on the Azotobacter vinelandii dataset. Particularly, the 10 noisy data matrices are obtained by randomly shuffling the nonzero entries of each row of two valid matrices, which are constructed in the same way as in the first type of experiments. For reference, we also applied MIMLNN on the same dataset with the same 10 noisy data matrices, and reported the results in Fig. \ref{fig5a}.
\begin{figure}[h!t]
\centering
\subfigure[\emph{1-RankLoss}]{\label{fig5a}\includegraphics[scale=0.23]{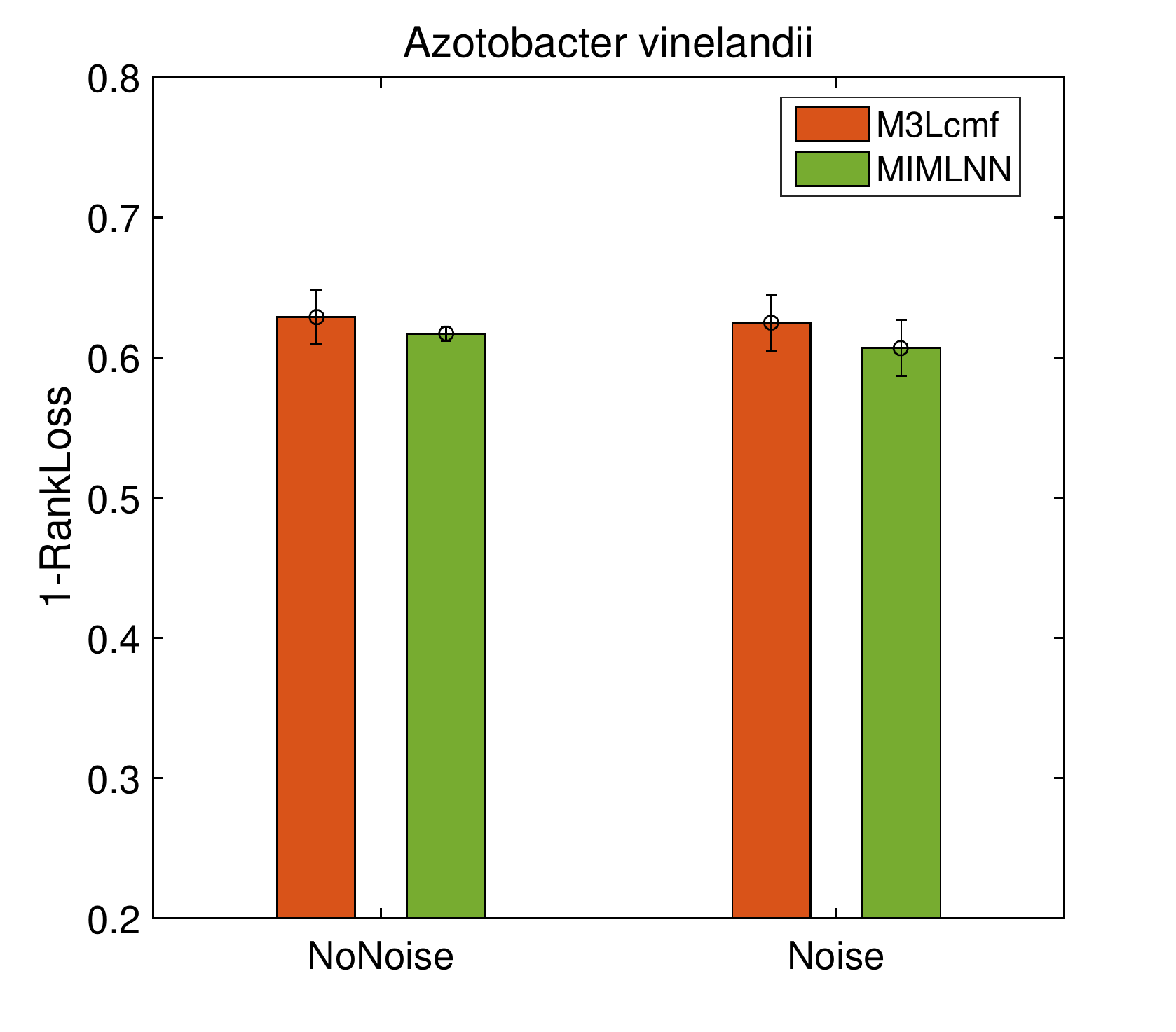}}
\subfigure[\emph{Weights}]{\label{fig5b}\includegraphics[scale=0.23]{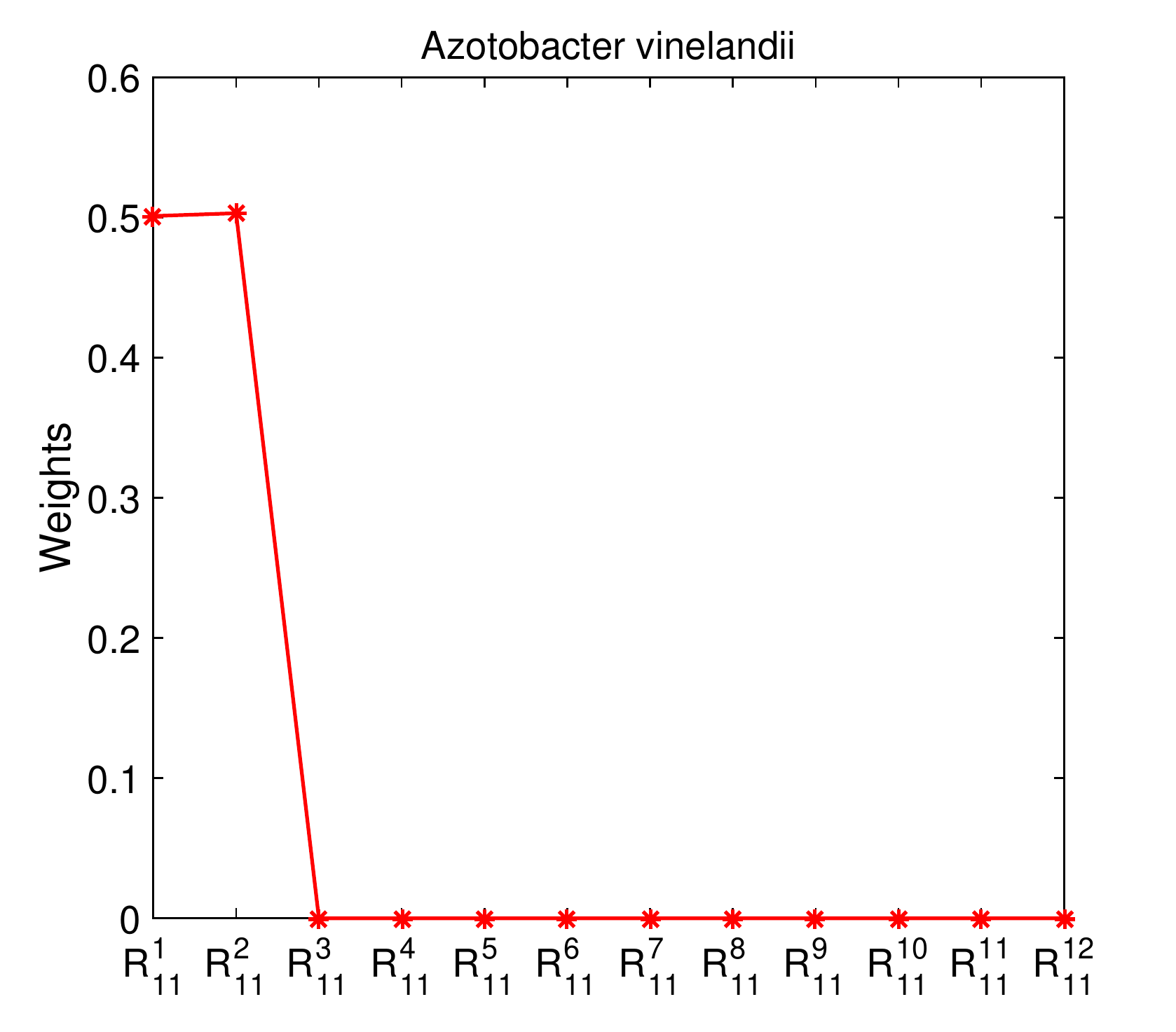}}
 \vspace{-0.5em}
 \caption{(a) Prediction results with and without the noisy data matrices, (b) Weights assigned by M3Lcmf to 12 intra-relational data matrices of bags. The first 2 are valid data matrices, and the last 10 are noisy ones.}
 \vspace{-0.5em}
 \label{fig5}
\end{figure}

Even with 10 noisy data matrices, M3Lcmf does not show a decreased performance, but MIMLNN shows a clearly reduced performance (by 2\%). That is because M3Lcmf explicitly considers the different relevances of intra-relational data matrices, and it can selectively integrate these matrices. In contrast, MIMLNN does not account for the different relevances of these matrices. As a result, it is more impacted by these noisy matrices.

To further investigate the underlying reason for the robust performance of M3Lcmf, we plot weights assigned to these 12 (2 valid and 10 noisy) intra-relational data matrices of bags in Fig. \ref{fig5b}. We can see that these 10 noisy data matrices are assigned with zero weights. Namely, M3Lcmf discards these noisy data matrices during the collaborative matrix factorization process. This investigation justifies our motivation to account for different relevances of multiple intra-relation data matrices.
\section{Conclusion}
In this paper, we proposed a collaborative matrix factorization based multi-view multi-instance multi-label learning approach called M3Lcmf. M3Lcmf utilizes a heterogeneous network to capture different types of relations in M3L, and collaboratively factorizes the relational data matrices of the network to explore the intrinsic relations between bags, instances, and labels. Extensive experimental results on different datasets corroborate our hypothesis that multiple types of relations can boost the performance of M3L, and their joint usage contributes to a significantly improved performance of M3Lcmf against  competitive approaches. The Supplementary file and code of M3Lcmf are available at \url{http://mlda.swu.edu.cn/codes.php?name=M3Lcmf}.

\section{ Acknowledgments}
The authors appreciate the reviewers for their helpful comments on improving our work. This work is supported by NSFC (61872300, 61741217, 61873214 and 61871020), NSF of CQ CSTC (cstc2018jcyjAX0228, cstc2016jcyjA0351 and CSTC2016SHMSZX0824), the Open Research Project
of Hubei Key Laboratory of Intelligent Geo-Information Processing (KLIGIP-2017A05), the National Science and Technology Support Program
(2015BAK41B03 and 2015BAK41B04), and Fundamental Research Funds for the Central Universities of China (XDJK2019D019 and XDJK2019B024).

\bibliographystyle{aaai}
\bibliography{M3L_Bib}

\begin{thebibliography}{}

\bibitem[\protect\citeauthoryear{Belkin, Niyogi, and
  Sindhwani}{2006}]{belkin2006manifold}
Belkin, M.; Niyogi, P.; and Sindhwani, V.
\newblock 2006.
\newblock Manifold regularization: A geometric framework for learning from
  labeled and unlabeled examples.
\newblock {\em JMLR} 7(11):2399--2434.

\bibitem[\protect\citeauthoryear{Blei, Ng, and Jordan}{2003}]{Blei2003Latent}
Blei, D.~M.; Ng, A.~Y.; and Jordan, M.~I.
\newblock 2003.
\newblock Latent dirichlet allocation.
\newblock {\em Journal of Machine Learning Research} 3:993--1022.

\bibitem[\protect\citeauthoryear{Briggs, Fern, and
  Raich}{2012}]{briggs2012rank}
Briggs, F.; Fern, X.~Z.; and Raich, R.
\newblock 2012.
\newblock Rank-loss support instance machines for miml instance annotation.
\newblock In {\em KDD},  534--542.

\bibitem[\protect\citeauthoryear{Chen \bgroup et al\mbox.\egroup
  }{2018}]{chen2018CMAL}
Chen, X.; Yu, G.; Domeniconi, C.; Wang, J.; Li, Z.; and Zhang, Z.
\newblock 2018.
\newblock Cost effective multi-label active learning via querying subexamples.
\newblock In {\em ICDM},  1--6.

\bibitem[\protect\citeauthoryear{Dem{\v{s}}ar}{2006}]{demvsar2006statistical}
Dem{\v{s}}ar, J.
\newblock 2006.
\newblock Statistical comparisons of classifiers over multiple data sets.
\newblock {\em JMLR} 7(1):1--30.

\bibitem[\protect\citeauthoryear{Feng and Zhou}{2017}]{feng2017deep}
Feng, J., and Zhou, Z.-H.
\newblock 2017.
\newblock Deep miml network.
\newblock In {\em AAAI},  1884--1890.

\bibitem[\protect\citeauthoryear{Gibaja and
  Ventura}{2015}]{gibaja2015mltutorial}
Gibaja, E., and Ventura, S.
\newblock 2015.
\newblock A tutorial on multilabel learning.
\newblock {\em ACM Computing Surveys} 47(3):52.

\bibitem[\protect\citeauthoryear{Gligorijevi{\'c} and
  Pr{\v{z}}ulj}{2015}]{gligorijevic2015methods}
Gligorijevi{\'c}, V., and Pr{\v{z}}ulj, N.
\newblock 2015.
\newblock Methods for biological data integration: perspectives and challenges.
\newblock {\em Journal of the Royal Society Interface} 12(112):20150571.

\bibitem[\protect\citeauthoryear{He \bgroup et al\mbox.\egroup
  }{2016}]{he2016online}
He, J.; Du, C.; Zhuang, F.; Yin, X.; He, Q.; and Long, G.
\newblock 2016.
\newblock Online bayesian max-margin subspace multi-view learning.
\newblock In {\em IJCAI},  1555--1561.

\bibitem[\protect\citeauthoryear{Huang, Gao, and Chen}{2017}]{huang2017multi}
Huang, S.-J.; Gao, N.; and Chen, S.
\newblock 2017.
\newblock Multi-instance multi-label active learning.
\newblock In {\em IJCAI},  1886--1892.

\bibitem[\protect\citeauthoryear{Huang, Gao, and Zhou}{2018}]{Huang2013Fast}
Huang, S.-J.; Gao, W.; and Zhou, Z.-H.
\newblock 2018.
\newblock Fast multi-instance multi-label learning.
\newblock {\em TPAMI} 99(1):1--14.

\bibitem[\protect\citeauthoryear{Lee and Seung}{2001}]{lee2001NMF}
Lee, D.~D., and Seung, H.~S.
\newblock 2001.
\newblock Algorithms for non-negative matrix factorization.
\newblock In {\em NIPS},  556--562.

\bibitem[\protect\citeauthoryear{Li \bgroup et al\mbox.\egroup
  }{2017}]{Li2017Multi}
Li, B.; Yuan, C.; Xiong, W.; Hu, W.; Peng, H.; Ding, X.; and Maybank, S.
\newblock 2017.
\newblock Multi-view multi-instance learning based on joint sparse
  representation and multi-view dictionary learning.
\newblock {\em TPAMI} 39(12):2554--2560.

\bibitem[\protect\citeauthoryear{Nguyen \bgroup et al\mbox.\egroup
  }{2014}]{Nguyen2014Labeling}
Nguyen, C.~T.; Wang, X.; Liu, J.; and Zhou, Z.~H.
\newblock 2014.
\newblock Labeling complicated objects: multi-view multi-instance multi-label
  learning.
\newblock In {\em AAAI},  2013--2019.

\bibitem[\protect\citeauthoryear{Nguyen, Zhan, and
  Zhou}{2013}]{Nguyen2013Multi}
Nguyen, C.~T.; Zhan, D.~C.; and Zhou, Z.~H.
\newblock 2013.
\newblock Multi-modal image annotation with multi-instance multi-label lda.
\newblock In {\em IJCAI},  1558--1564.

\bibitem[\protect\citeauthoryear{Shao \bgroup et al\mbox.\egroup
  }{2016}]{shao2016multi}
Shao, W.; Zhang, J.; He, L.; and Philip, S.~Y.
\newblock 2016.
\newblock Multi-source multi-view clustering via discrepancy penalty.
\newblock In {\em IJCNN},  2714--2721.

\bibitem[\protect\citeauthoryear{Tan \bgroup et al\mbox.\egroup
  }{2018}]{tan2018incomplete}
Tan, Q.; Yu, G.; Domeniconi, C.; Wang, J.; and Zhang, Z.
\newblock 2018.
\newblock Incomplete multi-view weak-label learning.
\newblock In {\em IJCAI},  2703--2709.

\bibitem[\protect\citeauthoryear{Villani}{2008}]{villani2008optimal}
Villani, C.
\newblock 2008.
\newblock {\em Optimal transport: old and new}, volume 338.
\newblock Springer Science \& Business Media.

\bibitem[\protect\citeauthoryear{Winn, Criminisi, and
  Minka}{2005}]{winn2005object}
Winn, J.; Criminisi, A.; and Minka, T.
\newblock 2005.
\newblock Object categorization by learned universal visual dictionary.
\newblock In {\em ICCV},  1800--1807.

\bibitem[\protect\citeauthoryear{Xu, Tao, and Xu}{2013}]{xu2013survey}
Xu, C.; Tao, D.; and Xu, C.
\newblock 2013.
\newblock A survey on multi-view learning.
\newblock {\em arXiv preprint arXiv:1304.5634}.

\bibitem[\protect\citeauthoryear{Yang \bgroup et al\mbox.\egroup
  }{2018}]{Yang2018KDD}
Yang, Y.; Wu, Y.-F.; Zhan, D.-C.; Liu, Z.-B.; and Jiang, Y.
\newblock 2018.
\newblock Complex object classification: A multi-modal multi-instance
  multi-label deep network with optimal transport.
\newblock In {\em KDD},  2594--2603.

\bibitem[\protect\citeauthoryear{Zhang and Wang}{2009}]{Zhang2009MIMLRBF}
Zhang, M.~L., and Wang, Z.~J.
\newblock 2009.
\newblock Mimlrbf: Rbf neural networks for multi-instance multi-label learning.
\newblock {\em Neurocomputing} 72(16-18):3951--3956.

\bibitem[\protect\citeauthoryear{Zhang and Zhou}{2009}]{Zhang2009}
Zhang, M.-L., and Zhou, Z.-H.
\newblock 2009.
\newblock Multi-instance clustering with applications to multi-instance
  prediction.
\newblock {\em Applied Intelligence} 31(1):47--68.

\bibitem[\protect\citeauthoryear{Zhang and Zhou}{2014}]{zhang2014mlreview}
Zhang, M., and Zhou, Z.
\newblock 2014.
\newblock A review on multi-label learning algorithms.
\newblock {\em TKDE} 26(8):1819--1837.

\bibitem[\protect\citeauthoryear{Zhou \bgroup et al\mbox.\egroup
  }{2008}]{Zhou2008MIML}
Zhou, Z.~H.; Zhang, M.~L.; Huang, S.~J.; and Li, Y.~F.
\newblock 2008.
\newblock Miml: A framework for learning with ambiguous objects.
\newblock {\em Corr Abs}  2012.

\bibitem[\protect\citeauthoryear{Zhou \bgroup et al\mbox.\egroup
  }{2012}]{Zhou2012MIML}
Zhou, Z.-H.; Zhang, M.-L.; Huang, S.-J.; and Li, Y.-F.
\newblock 2012.
\newblock Multi-instance multi-label learning.
\newblock {\em Artificial Intelligence} 176(1):2291--2320.

\bibitem[\protect\citeauthoryear{Zhu, Ting, and Zhou}{2017}]{zhu2017discover}
Zhu, Y.; Ting, K.~M.; and Zhou, Z.-H.
\newblock 2017.
\newblock Discover multiple novel labels in multi-instance multi-label
  learning.
\newblock In {\em AAAI},  2977--2984.

\bibitem[\protect\citeauthoryear{Zitnik and Zupan}{2015}]{2015Zitnik}
Zitnik, M., and Zupan, B.
\newblock 2015.
\newblock Data fusion by matrix factorization.
\newblock {\em TPAMI} 37(1):41--53.

\end{thebibliography}
\end{document}